# TRANSFORMER MODEL DETECTS ANTIDEPRESSANT USE FROM A SINGLE NIGHT OF SLEEP, UNLOCKING AN ADHERENCE BIOMARKER

Ali Mirzazadeh[1], Simon Cadavid[2], Kaiwen Zha[1], Chao Li[1], Sultan Alzahrani[3], Manar Alawajy[3], Joshua Korzenik[4], Kreshnik Hoti[5], Charles Reynolds[6], David Mischoulon[7], John Winkelman[7], Maurizio Fava[7], and Dina Katabi[1]

[1] Massachusetts Institute of Technology (MIT), Cambridge, MA, USA
[2] University of Michigan, Ann Arbor, MI, USA
[3] King Abdulaziz City for Science and Technology (KACST), Riyadh, Saudi Arabia
[4] Brigham & Women's Hospital & Harvard Medical School, Boston, MA, USA
[5] University of Prishtina, Faculty of Medicine, Division of Pharmacy, Kosovo
[6] University of Pittsburgh Medical Center, Pittsburgh, PA, USA
[7] Massachusetts General Hospital & Harvard Medical School, Boston, MA, USA

One Sentence Summary:  We introduce a transformer-based model that detects antidepressant use from a single night of contact-free, in-home sleep monitoring, opening the door to scalable remote tracking of medication adherence in mood disorders.

# ABSTRACT

Antidepressant nonadherence is pervasive, driving relapse, hospitalization, suicide risk, and billions in avoidable costs. Clinicians need tools that detect adherence lapses promptly, yet current methods are either invasive (serum assays, neuroimaging) or proxy-based and inaccurate (pill counts, pharmacy refills). We present the first noninvasive biomarker that detects antidepressant intake from a single night of sleep. A transformer-based model analyzes sleep data from a consumer wearable or contactless wireless sensor to infer antidepressant intake, enabling remote, effortless, daily adherence assessment at home. Across six datasets comprising 62,000 nights from >20,000 participants (1,800 antidepressant users), the biomarker achieved AUROC = 0.84, generalized across drug classes, scaled with dose, and remained robust to concomitant psychotropics. Longitudinal monitoring captured real-world initiation, tapering, and lapses. This approach offers objective, scalable adherence surveillance with potential to improve depression care and outcomes.

# INTRODUCTION

Antidepressant non-adherence is a pervasive obstacle to the effective treatment of depressive disorders[1]. Real-world data indicate that roughly half of patients with major depressive disorder (MDD) are already non-adherent just six months after initiating therapy[2]. Such adherence lapses are associated with over 30% more hospitalizations, higher emergency-department utilization, and an additional $500–$1,600 in annual medical costs per patient compared with adherent peers[3] burdening the healthcare system by billions of dollars. A 2022 meta-analysis of 40 randomized trials showed that continuing antidepressants after remission lowers relapse rates by about 24% relative to discontinuation[4], highlighting how even brief interruptions increase the risk of depressive recurrence. The most alarming downstream consequence of nonadherence is suicidality. In a U.S. cohort of 2.4 million patients, prematurely stopping an antidepressant heightened the odds of a suicide attempt by 61% [5]; Similarly, U.K. data revealed peaks in self-harm and suicide during the first month after discontinuation.[6]

These findings underscore an urgent clinical and economic need for reliable antidepressant-adherence monitoring to detect and address non-adherence in a timely manner. Yet existing methods are ill-suited for routine care. Objective measures such as serum drug assays[7][8][9], electroencephalography (EEG)[10], and neuroimaging[11][12] can confirm drug ingestion but require clinic visits and are prohibitively expensive for day-to-day management. Scalable alternatives such as mobile-health apps, self-report questionnaires, and pill counts rely on surrogates that do not measure drug intake and have been shown inaccurate[13][14]. Most also hinge on active patient engagement, which is problematic in depression, where reduced motivation, cognitive deficits, and executive dysfunction are common[15][16]. Patients who struggle to take their medication often face the same obstacles when asked to self-monitor, and many conceal lapses out of embarrassment or stigma[17]. Pharmacy claims and refill records[18] are less subjective but provide only delayed, coarse feedback and similarly reflect prescription refill rather than medication intake[19]. Consequently, an accurate and efficient antidepressant-adherence monitoring solution remains an unmet public-health priority.

Our study presents a cost-effective, fully remote solution that enables clinicians to verify antidepressant intake without clinic visits or patient input. The method harnesses the influence of antidepressants on sleep and EEG physiology, e.g., REM suppression and alterations in slow-wave and beta activity[20][10], and uses artificial intelligence (AI) to translate these effects from population level findings[21] into an objective biomarker suitable for individual-level routine monitoring (i.e., n=1).

Our solution detects antidepressant use from a single night of in-home sleep monitoring (Fig. 1). It takes as input nocturnal respiratory waveforms collected during sleep and uses AI to reconstruct an auxiliary EEG representation which it then leverages to classify medication status (as detailed in the "AI Model" section). Respiratory waveforms can be readily captured by consumer wearables[22][23] or contact-free radar-based systems such as Emerald[24][25], which captures respiration and sleep hypnograms without any on-body sensor[26], achieving performance comparable to polysomnography[27]. Adding medication adherence detection to such contact-free sensors enables effortless day-to-day monitoring, even in the absence of patient engagement. Further, such solution reflects the actual drug administration rather than indirect proxies like pill-bottle openings or pharmacy records.

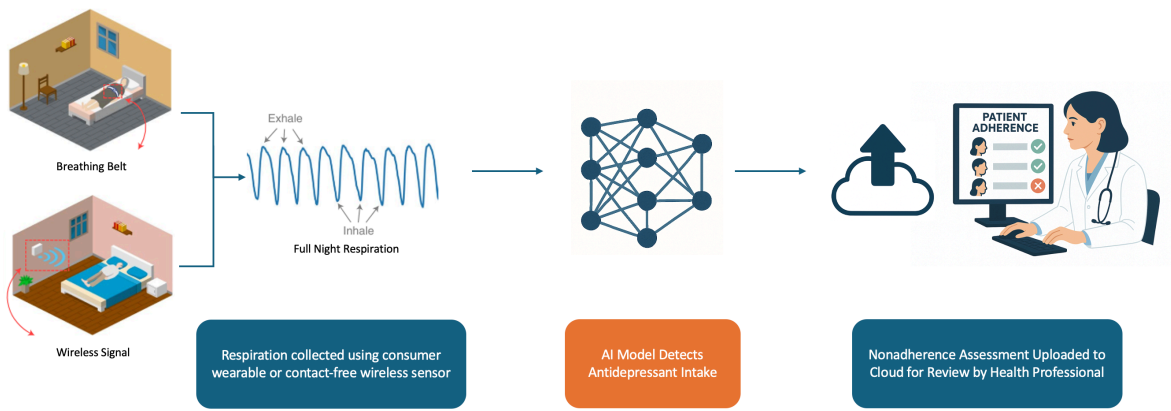

**Figure 1| Remote AI-powered antidepressant adherence monitoring**. *Nocturnal breathing signals from a wearable or contactless consumer sensor are analyzed by an AI model to detect antidepressant intake, with results delivered remotely to clinicians to support adherence assessment, dose adjustment, and care plan.*

We retrospectively validated our solution on a large multi-center sleep and respiratory dataset comprising 62,000 nights from over 20,000 individuals. The AI model achieved an AUROC of 0.84 across 11 different antidepressant agents, demonstrated dose-responsive scoring, and remained robust to potential medication-related confounders with central nervous system activity, including hypnotics, anxiolytics, anticholinergics, antipsychotics, and anticonvulsants. Longitudinal case studies highlighted its real-world utility, capturing antidepressant initiation, tapering, and adherence lapses, thereby underscoring its value for ongoing clinical oversight.

By enabling continuous, home-based monitoring of antidepressant adherence, this AI-driven system could help clinicians detect unexpected discontinuation early, which reduces relapse, hospitalization, and suicide risk, improves patient outcomes, and lowers healthcare costs.

# RESULTS

## Datasets and Participants

For our analysis, we curated 5 publicly available sleep datasets, Sleep Heart Health Study (SHHS), MrOS Sleep Study (MrOS), Wisconsin Sleep Cohort (WSC), Hispanic Community Health Study (HCHS), and Cleveland Family Study (CFS), along with our dataset from the Massachusetts Institute of Technology (MIT). While none of these datasets is focused on MDD or antidepressant use, they were chosen because they contain medication records with antidepressant labels. In total, the datasets contain 62,606 nights of sleep from 20,552 individuals, out of which 1,842 reported taking antidepressants (mean age: 60.5 ± 12.9 years, 59.6% women) and 18,710 are control subjects (mean age: 57.02 ± 16.2 years, 47.0% women). Table 1 summarizes the datasets characteristics.

The datasets can be divided into two categories:

1. Polysomnography (PSG) sleep studies that include respiratory signals, sleep staging, and electroencephalography (EEG) recordings, among others. The respiratory signals were collected using wearable belts that continuously monitored nocturnal breathing patterns. All PSG datasets used in this paper are publicly available, de-identified, and originally collected under protocols approved by their respective institutional review boards.

2. Wireless Monitoring Data – Obtained using a contactless wireless sensor that uses passive radio waves to track chest movements (the Emerald sensor). The dataset was collected under approved Research Review Boards (RSRB00001787, RSRB00003703, 2020P000348) and informed consent was obtained from all participants. Prior studies have demonstrated the high accuracy of these sensors, with the resulting respiratory signals achieving 92% average correlation with gold-standard respiratory effort belts. The sensors also achieved 80.5% accuracy for 4-class sleep stage classification and an inter-correlation coefficient of 0.90 for estimating the Apnea-Hypopnea Index [27][26].

| | Dataset | Cohort | Data Type | Source of Breathing Signal | No. of Nights (Antidepressant) | No. of Participants (Antidepressant) | No. of Nights per Participant (σ) | Sex (%) Female | Age in years - mean (σ) | Race (%) Asian | Black | White | Other |
|---|---|---|---|---|---|---|---|---|---|---|---|---|---|
| Internal | SHHS | Sleep Disorders & Cardiovascular Outcomes | PSG sleep study | Breathing Belt (Abdominal) | 8418 (703) | 5782 (585) | 1.46 (0.50) | 52.4 | 64.4 (11.2) | 0.0 | 8.8 | 84.7 | 6.5 |
| | MrOS | Sleep Disorders, Fractures, and Vascular Disease | PSG sleep study | Breathing Belt (Abdominal) | 3551 (294) | 2711 (255) | 1.31 (0.46) | 0.0 | 77.5 (5.48) | 3.1 | 3.5 | 90.8 | 2.7 |
| | HCHS | Hispanic Community Health Study / Study of Latinos | PSG sleep study | Airflow (Nasal Cannula) | 9754 (608) | 9754 (608) | 1.0 (0) | 58.5 | 47.4 (12.9) | 0.3 | 2.6 | 35.5 | 58.8 |
| | CFS | Sleep Apnea, Family-based | PSG sleep study | Breathing Belt (Abdominal) | 499 (37) | 499 (37) | 1.0 (0) | 56.1 | 47.2 (15.8) | 0.0 | 55.5 | 42.5 | 2.0 |
| | MIT | varies | Sleep Study (at-home) | Wireless | 37815 (10503) | 145 (42) | 260.79 (158.34) | 51.7 | 47.0 (18.01) | 4.8 | 2.1 | 57.9 | 35.2 |
| External | WSC | Longitudinal Study of Sleep Disorders | PSG sleep study | Breathing Belt (Abdominal) | 2569 (541) | 1122 (315) | 2.29 (1.04) | 45.9 | 59.3 (8.0) | 1.1 | 2.0 | 94.8 | 2.0 |
| | Overall | | | | 62606 (12686) | 20013 (1842) | 3.13 (25.82) | 48.0 | 57.2 (16.0) | 0.7 | 5.8 | 60.9 | 31.3 |

**Table 1 | Overview of datasets used for training and evaluation.** *The study combined six datasets: five public polysomnography datasets, each contributing between one and five nights per participant, and one internal dataset from MIT consisting of in-home wireless respiration recordings with longitudinal monitoring over hundreds of nights per individual. In total, more than 62,000 nights of sleep data from 20,000 patients were used for model development and analysis.*

**Antidepressant Status:** Medication status was determined from the medication records in each dataset and reflected recent medication intake, as detailed in Supplemental Material. Reporting granularity differed: some datasets listed individual drug names, whereas others noted only therapeutic classes: selective serotonin-reuptake inhibitors (SSRIs), serotonin-norepinephrine reuptake inhibitors (SNRIs), tricyclic antidepressants (TCAs), or "other."  Trazodone was excluded from the analysis because it is often prescribed at low, sedative doses to treat insomnia or anxiety, and the datasets lacked dose-specific annotations. The MIT dataset provided longitudinal coverage (1–2 years per participant) with detailed start/stop dates and dose information for each antidepressant; therefore, we relied on this cohort to analyze dose effects and to capture initiation, tapering, and discontinuation events.

Additional details about sampling rates, labels, and quality control are provided in the supplemental material.

# AI Model

We developed an AI model that detects antidepressant exposure from nocturnal respiration, using EEG reconstruction as an auxiliary objective to guide representation learning. We deliberately did not use EEG as the model input because obtaining sleep EEG at home is burdensome requiring overnight electrodes on the head, which would limit usability for routine adherence monitoring.

The model takes as input one night of respiratory signal and outputs a classification score z in [0,1], where z=1 denotes detection of antidepressant use (Fig. 2). The model has three components: an encoder, a decoder, and a classification module. The encoder and decoder were jointly pre-trained with an auxiliary EEG-reconstruction task. After pretraining, the encoder was frozen, and a classifier was trained to predict antidepressant use from the learned EEG representation. The encoder is a transformer with 8 blocks, 8 self-attention heads per block, and a 768-dimensional embedding, configured as a ViT-Small architecture[28]. The decoder mirrors the encoder with 8 transformer blocks and is trained to reconstruct EEG spectrograms from the encoder's latent representations.

The encoder outputs are concatenated with learned positional embeddings and passed to the antidepressant classifier. The classifier is a transformer composed of four self-attention layers, each with a hidden dimension and four attention heads. A learnable classification token is prepended to the input sequence, and its final embedding is used by a classification head for binary prediction. Critically, the model operates solely on respiration at inference and does not need access to EEG data.

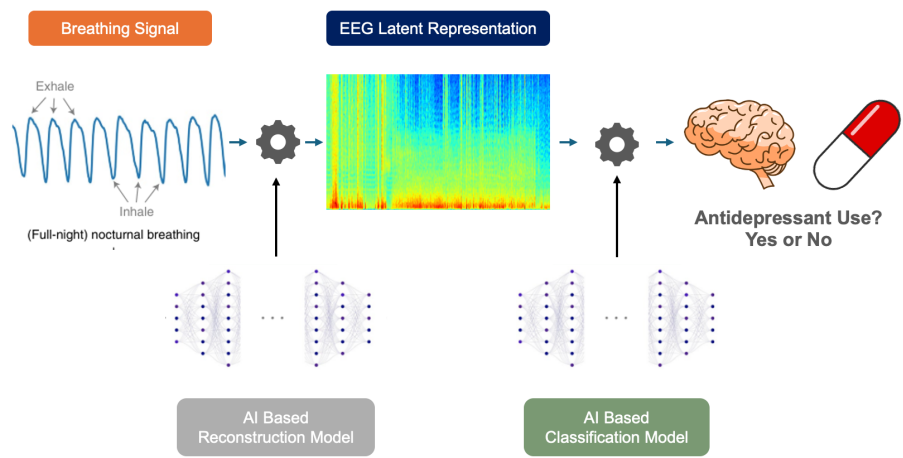

**Figure 2 | Overview of the antidepressant detection model.** *Nocturnal respiration signals collected from a wearable or contactless wireless sensor are input into an encoder model to generate an auxiliary EEG representation, which is passed to a classification model that estimates the likelihood of antidepressant use.*

**Model Training:** A 4-fold stratified cross-validation scheme was applied to all datasets except for the Wisconsin Sleep Cohort (WSC), which was held out entirely for external evaluation. Stratification was based on antidepressant use. All stages of model training used the same 4-fold splits to ensure consistency and prevent information leakage. Individuals assigned to training folds were strictly excluded from test sets throughout the entire pipeline.

Models were trained using the AdamW optimizer. To address class imbalance, the antidepressant-positive class was oversampled within each training fold. Training proceeded for 4,000 steps per fold using a batch size of 48. A cosine annealing learning rate scheduler with linear warmup was employed to stabilize optimization. Model checkpoints were saved from the final training step, without early stopping or validation-based tuning.

## Statistical Analysis and Performance Measures

Group comparisons of sleep features (e.g., antidepressant vs. control) were conducted using Welch's t-test, which accommodates unequal variances. Effect sizes were reported as the absolute value of Cohen's d, computed as the difference in group means divided by the square root of the average of the group variances.

AI Model performance was assessed by the area under the receiver operating characteristic curve (AUROC), averaged across predictions from all four folds. To account for unequal numbers of nights per participant, nightly prediction scores from the same individual were averaged; when a participant initiated or discontinued an antidepressant, scores from each phase were averaged separately. AUROC values are reported with 95% confidence intervals derived from 1,000 bootstrap resamples of the full evaluation dataset. Sensitivity and specificity were computed at a single operating point defined by the threshold that maximized Youden's J statistic on the pooled validation set. To address class imbalance, positive predictive value (PPV) and negative predictive value (NPV) were calculated under a balanced class ratio obtained by randomly subsampling the negative class to match the positive class; PPV and NPV are accompanied by 95% confidence intervals from 1,000 bootstrap iterations. Analyses of medication, dosage, and confounds were limited to antidepressant monotherapy users to minimize confounding.

For comparative evaluation, two random-forest classifiers were trained using sleep features previously associated with antidepressant use[29,21,30]. Random forests were implemented with 1,000 trees and a maximum depth of 10; all other parameters followed scikit-learn defaults.

## Capturing Antidepressant Impact on Sleep Architecture via Nocturnal Respiration

Because our model uses only respiration, we first investigated whether respiration-derived sleep staging can recover established antidepressant effects on sleep. Using the 5 PSG datasets (N = 12,043; 1,328 on antidepressants), we compared expert-annotated hypnograms with hypnograms generated by a published AI model that analyzes respiration alone to perform sleep staging[27]. Across all participants (Fig. 3a), both expert-labeled and respiration-based hypnograms showed significant prolongation of REM latency and reductions in REM duration in the antidepressant cohort (all p < 1e-10), with the largest effect for REM latency (Cohen's d = [0.88, 0.87]). Sleep Efficiency increased in the antidepressant group (p < 1e-10), while slow-wave sleep (SWS) was reduced (p < 0.001). Importantly, the expert-labeled and respiration-based hypnograms revealed the same inter-cohort patterns. Pairwise comparisons of features derived from the two methods (Fig. 3b) showed high correlations for REM latency, REM duration, and sleep efficiency in both control and antidepressant groups (r = [0.69-0.87]); SWS duration correlated moderately in controls and weakly in the antidepressant group (r = [0.40, 0.37]). These results confirm that sleep stages extracted from nocturnal breathing using AI capture the impact of antidepressant observed in PSG sleep data, supporting our use of nocturnal breathings as the input data modality to our AI.

Next, we zoomed in on a few individuals to check whether antidepressant use is detectable at the individual level, as opposed to the population level. Specifically, visualization of REM episodes over time (Fig. 3c) demonstrated early-night REM suppression in individuals taking antidepressants. Collectively, these findings indicate that antidepressant effects on sleep, especially REM, provide a strong differentiator at the population level and a reliable within-person signal when aggregated over months of data, supporting their use for antidepressant medication adherence evaluation. However, for routine monitoring in clinical practice, an adherence biomarker must achieve sufficient accuracy at the individual level using one or a few nights of data. We next assess whether our AI model can deliver such biomarker.

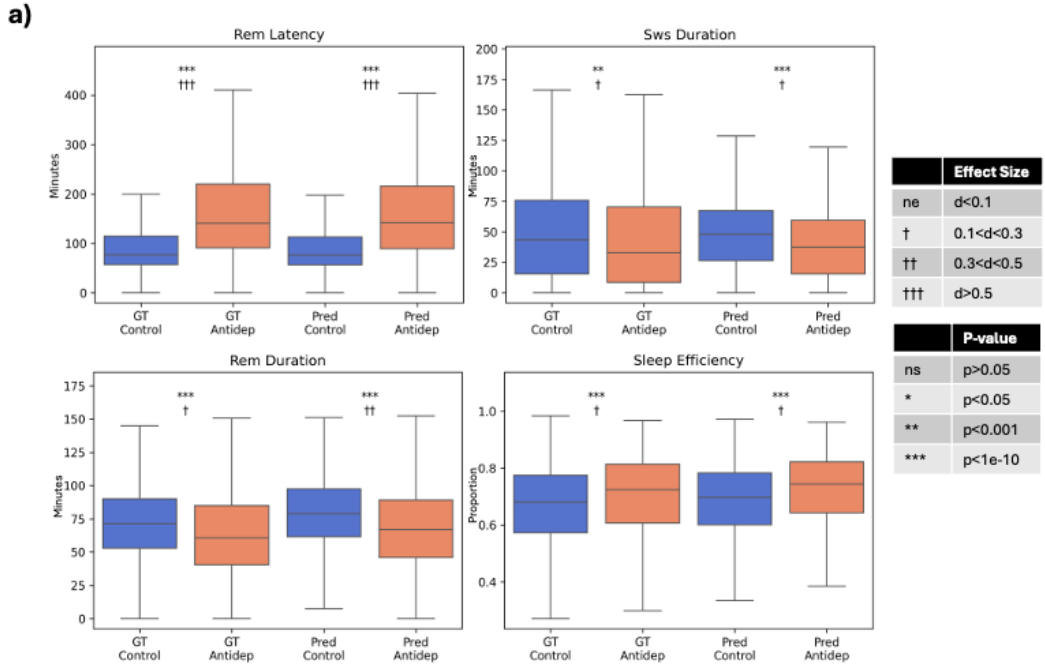

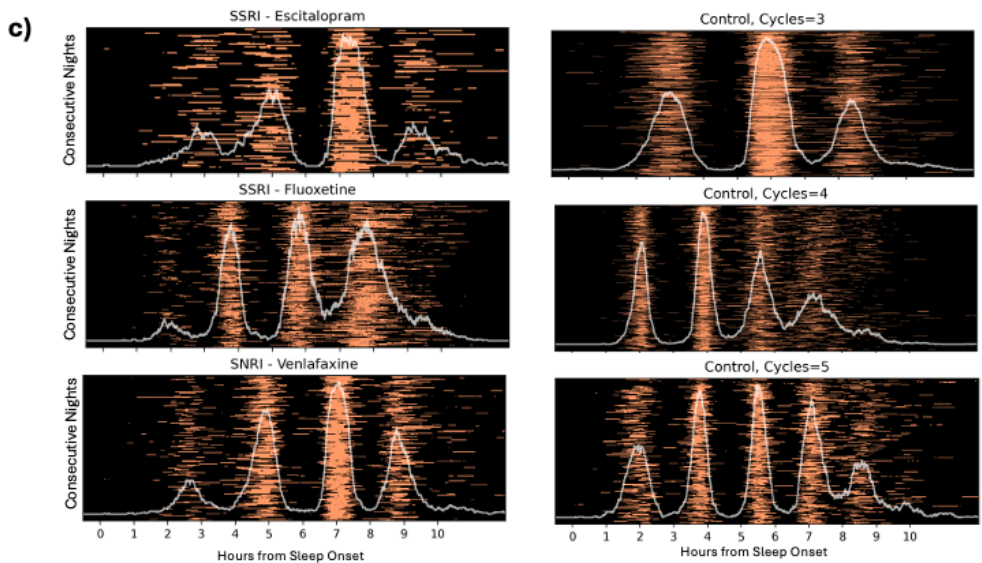

**Figure 3 | Analysis of antidepressant sleep features.** *a, Comparison of expert-labeled PSG-based and AI-generated respiration-based sleep hypnogram features between the antidepressant cohort and controls* ***b****, statistical*

*comparison of sleep features from expert-labeling and respiration reveals high correlation between the two sources, shown in both cohorts **c**, Longitudinal sleep hypnograms for six individuals with 3 antidepressant users (right column) and 3 control participants (left column). Each row in each sub-figure represents one night of sleep, with hours since sleep onset on the x-axis and consecutive nights on the y-axis. Bright segments indicate REM sleep; black indicates all other sleep stages. REM suppression can be consistently observed in the first sleep cycle of antidepressant users.*

## Antidepressant Detection from Single-Night Nocturnal Breathing

Assessment of the AI model showed that it accurately identified antidepressant use from single-night respiration recordings (Fig. 4a). Across datasets, the model achieved a mean AUROC of 0.84 ± 0.03. Further, the model maintained high accuracy when evaluated on the external holdout Wisconsin Sleep dataset (AUROC = 0.87), indicating robust generalization across populations and recording conditions.

For comparative evaluation, two random-forest classifiers were trained using sleep features previously associated with antidepressant use. The first incorporated standard sleep-stage and architecture features (e.g., stage durations, sleep and REM onset latencies, awakenings, and sleep efficiency) [29] [21] [30]. The second additionally included EEG-derived band-power features (SO, delta, theta, alpha, beta), computed both across the full night and within individual sleep stages[10]. (The detailed feature lists are provided in Supplemental Figure E). These models achieved AUROCs of 0.75 and 0.77, respectively. The superior performance of the respiration-based deep learning model suggests that it captures novel physiological information not reflected in traditional metrics.

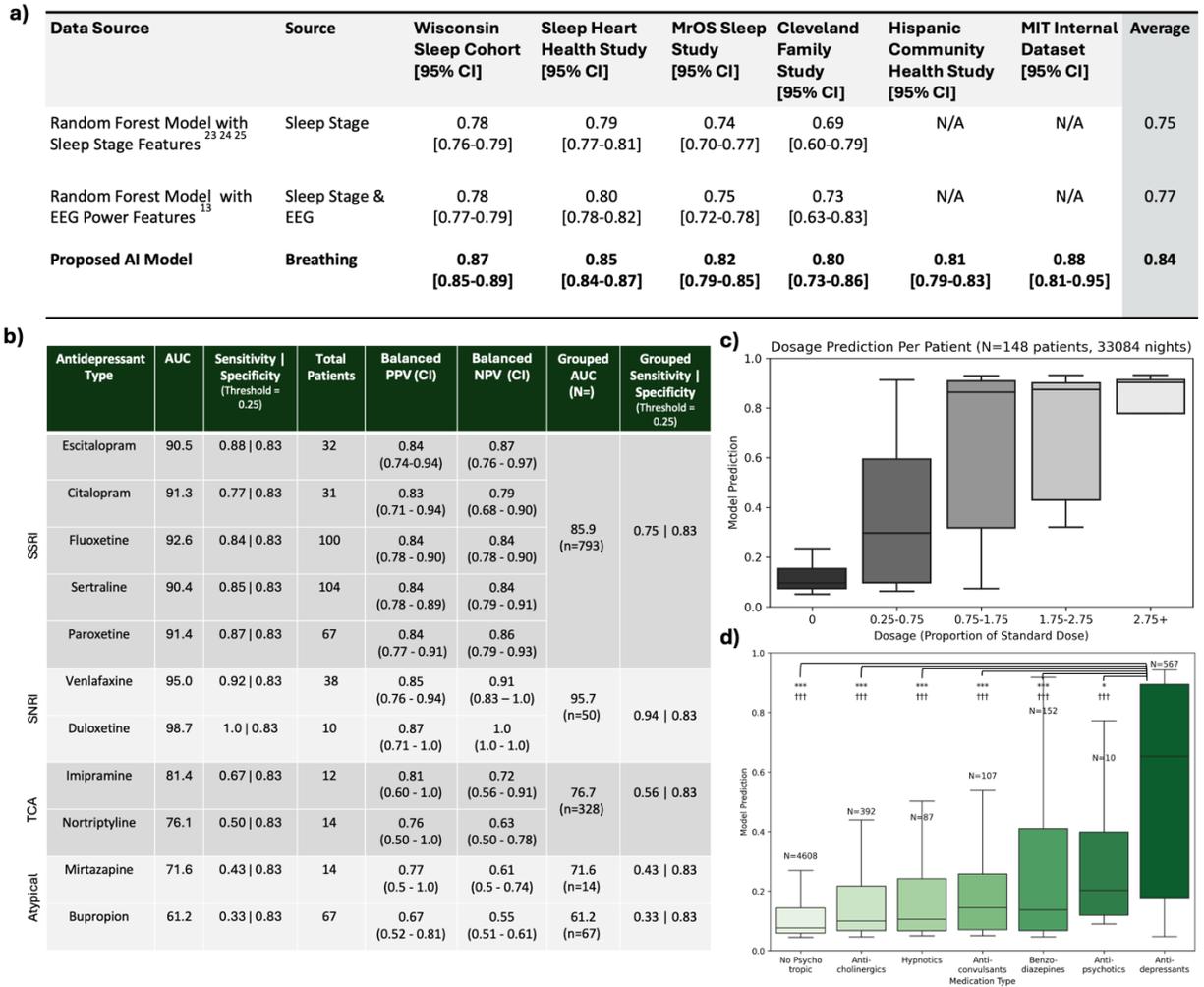

**Figure 4 | Evaluation of biomarker across datasets, antidepressant types, dosages, and other psychiatric medications.** *a, Model performance across datasets, reported as AUROC, with bootstrapped 95% confidence intervals. The biomarker significantly outperformed baseline models based on sleep stage features in all applicable datasets. b, Model performance stratified by antidepressant medication. AUROC values were relatively consistent for selective serotonin reuptake inhibitors (SSRIs, 0.86) and serotonin–norepinephrine reuptake inhibitors (SNRIs, 0.96), with more variable performance among tricyclic antidepressants (TCAs, 0.77). Detection accuracy was lower for mirtazapine (0.72) and bupropion (0.61). c, Relationship between model output score and antidepressant dosage, expressed as a proportion of the standard recommended dose for MDD. Higher doses were associated with higher prediction scores. d, Comparison of model scores across psychotropic medication classes and drugs that suppress REM sleep such as anticholinergics. The biomarker showed high specificity for antidepressants relative to other medications, revealing robustness to potential drug confounders.*

## Effect of Medication Type on Biomarker

To evaluate biomarker performance for specific antidepressant medications, we performed a stratified analysis using three datasets with detailed prescription information (Fig. 4b). Eleven antidepressants were examined, spanning across selective serotonin reuptake inhibitors (SSRIs), serotonin–norepinephrine reuptake inhibitors (SNRIs), tricyclic antidepressants (TCAs), and atypical antidepressants (Mirtazapine, Bupropion).

Across drug classes, the biomarker showed strong discrimination for serotonergic agents and weaker performance for atypicals (Fig. 4b). At the individual drug level, all SSRIs achieved AUROC ≈0.90–0.93 (escitalopram 0.905; citalopram 0.913; fluoxetine 0.926; sertraline 0.904; paroxetine 0.914), with sensitivities 0.77–0.88 at a fixed specificity of 0.83 and balanced PPV/NPV clustered around 0.80–0.87. SNRIs performed best (venlafaxine AUROC 0.95; duloxetine AUROC 0.99, n=10), yielding a grouped AUROC of 0.957 with sensitivity 0.94 at the same specificity, and balanced NPV up to 1.0 in the small duloxetine cohort. In contrast, TCAs were moderate (imipramine AUROC 0.81; nortriptyline 0.76; grouped sensitivity 0.56), and atypicals were lowest (mirtazapine AUROC 0.72; bupropion 0.61; sensitivities 0.43 and 0.33). Using conventional benchmarks for AUROC interpretation[31], these results span "excellent" performance for SNRIs and most SSRIs (AUROC ≥0.90) through "moderate acceptable" for TCAs and Mirtazapine (0.70–0.80) and "poor–acceptable" for bupropion (0.60-0.70).

The higher accuracy observed for SNRIs and SSRIs has plausible biological underpinnings. Serotonergic and noradrenergic systems directly modulate brainstem respiratory drive and upper-airway motor tone during sleep: serotonergic neurons contribute to chemoreception and eupneic ventilation, while noradrenergic activity stabilizes the airway—a relationship leveraged clinically by noradrenergic therapies that reduce obstructive events[32]. In contrast to TCAs, mirtazapine, and bupropion, which enhance noradrenergic signaling without blocking the serotonin transporter (SERT), SSRIs and SNRIs increase serotonin via SERT inhibition.

Furthermore, the pattern—SNRIs highest accuracy, then SSRIs, TCAs intermediate, and atypicals lowest—mirrors well-described drug effects on sleep architecture. The top AUROCs for SNRIs are consistent with their strong impact on sleep, particularly REM suppression. In particular, venlafaxine showed the best discrimination which is consistent with past trials where venlafaxine reduced REM sleep, total sleep time and sleep efficiency while increasing N1[33]. REM inhibition by venlafaxine is likely due to its ability to increase monoaminergic tone. Duloxetine, another SNRI, likewise suppresses REM and increases stage N3[34]. SSRIs act in the same direction but with more inter-individual variability, consistent with their slightly lower AUROCs and observational links to impaired oxygenation[35]. TCAs can suppress REM, yet their sleep effects are heterogeneous[36]; historically, the activating TCA protriptyline modestly improved oxygenation without consistently reducing apnea frequency, echoing the observed mid-range performance. In contrast, mirtazapine improves continuity and increases slow-wave sleep with little to no REM suppression[37], and two randomized, placebo-controlled trials in OSA found no apnea improvement[38], coherent with weaker discrimination. Finally, bupropion typically delays REM onset but does not reduce total REM[39] (often increasing REM density), yielding nights that are closest to baseline and near-chance classification. Together, the literature supports that the biomarker is primarily more sensitive to agents with serotonergic (especially agents with SERT involvement) and REM-suppressing pharmacology.

*Effect of Dose on Biomarker*

We next examined whether antidepressant dose levels influence detectability. As clinical practice often involves initiating treatment at lower doses and then titrating upward based on patient response[40 41], biomarker detectability may depend on whether a sufficient pharmacological threshold is reached during sleep. To investigate the impact of doses, prescribed doses were normalized by dividing each dose by the standard recommended dose for major depressive disorder (MDD). Supplemental Figure D provides further details on dosage normalization. A graded association with dose was observed (Fig. 4c): classification performance was poorer for lower, potentially subtherapeutic doses and increased significantly at and above the standard dose. These findings indicate that the model is likely detecting physiological effects of antidepressants that become more pronounced with higher drug concentrations, and therefore increased serotonin reuptake inhibition, rather than merely identifying the presence of a prescribed medication.

*Assessment of Biomarker Specificity and Robustness to Medication Confounders*

To assess the biomarker's specificity for antidepressant detection, we compared model prediction scores in individuals taking antidepressants with those taking other classes of psychotropic medications (Fig. 4d), thereby evaluating potential confounding from these agents given their central nervous system activity and effects on sleep. We also assessed anticholinergic drugs, which, like antidepressants, suppress REM sleep and thus may confound detection. Moreover, many medications exhibit some anticholinergic activity; indeed, up to one half of medications prescribed in the elderly have this activity[42]. Accordingly, our aim here was to assess a wider spectrum of medications with potential confounding effects.

Our data shows that the biomarker scores for hypnotics, anticonvulsants, benzodiazepines, antipsychotics, and anticholinergics were all significantly lower than those for antidepressants ($p < 0.05$ for all comparisons, *t*-test), indicating that the biomarker is not confounded by the use of such medications.

The monotonic rise in the biomarker score, from no psychotropic (lowest), through anticholinergics, hypnotics and anticonvulsants, to benzodiazepines and antipsychotics, with antidepressants being highest , indicates that the biomarker is primarily sensitive to sleep stage composition (especially REM) and stage-linked breathing. The small increase in the biomarker score for antipsychotics drugs can be attributed to that first-generation antipsychotics (FGAs) generally increase REM latency, but their effects on sleep stage distribution are inconsistent across agents[43]. Second-generation antipsychotics (SGAs) more reliably increase total sleep time/sleep efficiency and often increase SWS, while REM effects are mixed—some drugs lengthen REM latency, others leave REM unchanged, and quetiapine can even reduce SWS and REM[44,45]. This heterogeneity means antipsychotics do not produce a single, stage-pattern "fingerprint," so a breathing-during-sleep biomarker that is most sensitive to consistent REM suppression/fragmentation (as with SNRIs/SSRIs) will rise only slightly on average for antipsychotics.

Benzodiazepines at therapeutic doses produce only modest, inconsistent reductions in REM but reliably increase N2 and decrease slow-wave sleep, with small increases in REM latency in some studies[46,47]. This can explain a small but more dispersed rise in score observed in the benzodiazepine group. By contrast, most Z-hypnotics produce relatively small, mixed effects on sleep stages (eszopiclone modestly prolongs REM latency[48]; zolpidem often preserves architecture[49]), so their distributions sit very close to the control baseline. Finally, anticonvulsants have a variable effect on sleep architecture. Classic anticonvulsants such as carbamazepine, which is studied more, suggest an increase in slow wave sleep and reduction of REM phase, whereas newer drugs, such as gabapentin in this category have an overall beneficial effect. Nonetheless, the evidence for this category of drugs is weak posing difficulties to establish related recommendations[50].

Supplemental Figure A extends these confounder analyses to drug co-therapy, assessing biomarker performance when multiple psychotropic drugs are combined.

*Effect of Sleep Apnea on Biomarker*

Since sleep apnea (SA) has been linked with exacerbation of depressive symptoms[51] and its severity also associated with higher depression risk[52], we assessed the biomarker performance across varying levels of sleep apnea severity in three major antidepressant classes. Data were restricted to cohorts with expert-labeled apnea events and apnea–hypopnea index (AHI) calculations. The AHI was determined following AASM clinical guidelines, requiring a >30% reduction in nasal cannula flow with ≥4% oxygen desaturation per hour of sleep[53]. Participants were classified as Normal (AHI < 5), Mild SA (AHI 5–15), Moderate SA (AHI 15–30), or Severe SA (AHI > 30). Model score distributions were then compared between antidepressant and control nights.

While SSRIs and SNRIs demonstrated non-significant changes to model score with increasing apnea severity, model score for TCA decreased in the presence of sleep apnea. This could be due to the sedative effects of TCAs being counteracted by the increased arousal associated with sleep apnea. Importantly, individuals on TCA with mild and moderate apnea maintained significantly higher model scores compared to the control cohort (p < 0.0001). More details can be accessed in Supplemental Figure B.

*Effect of Age and Sex on Biomarker*

To further evaluate biomarker performance, we stratified the dataset by age and sex and calculated area under the receiver operating characteristic curve (AUROC) values for each subgroup. Ninety-five percent confidence intervals were estimated by bootstrapping (N = 1000).

The model achieved its highest performance in the 50–60 age group (AUROC: 0.85 [0.82–0.87]) and its lowest in the 20–30 age group (AUROC: 0.74 [0.61–0.86]). The wider confidence intervals in the 20–30 group reflect greater variability, likely attributable to the smaller number of positive samples (only 1.4% of the total antidepressant cohort). Performance remained robust in older populations, including the 70–80 (AUROC: 0.82 [0.80–0.85]) and 80–90 (AUROC: 0.83 [0.79–0.87]) age groups. Stratification by sex indicated slightly higher performance in the female cohort compared with the male cohort (AUROC: 0.84 [0.82–0.85] vs. 0.82 [0.80–0.85]).

## Case Studies of Longitudinal Monitoring of Antidepressant Use

We evaluate the biomarker's utility for longitudinal monitoring by analyzing within-subject changes in scores over time. The MIT dataset included participants who experienced changes in antidepressant regimens during the recording period, providing a naturalistic setting to assess the model's sensitivity to real-world medication dynamics. We present (Fig. 5) four cases illustrating medication initiation, discontinuation, dose tapering, and variability in adherence (missed doses) across different antidepressant classes.

While antidepressant intake is a binary event, the model's longitudinal outputs suggest that the corresponding physiological signatures evolve gradually. In many cases, changes in model predictions stabilized only after several weeks, consistent with pharmacodynamic and pharmacokinetic principles requiring sustained medication exposure to elicit full neurophysiological adaptation. Although the datasets lack comprehensive adherence information or clinical outcome data, these case studies demonstrate the model's potential to provide continuous, quantitative insights into antidepressant use that is sensitive to potential fluctuations in medication blood plasma levels reflective of their administration, supporting its application for real-world treatment monitoring.

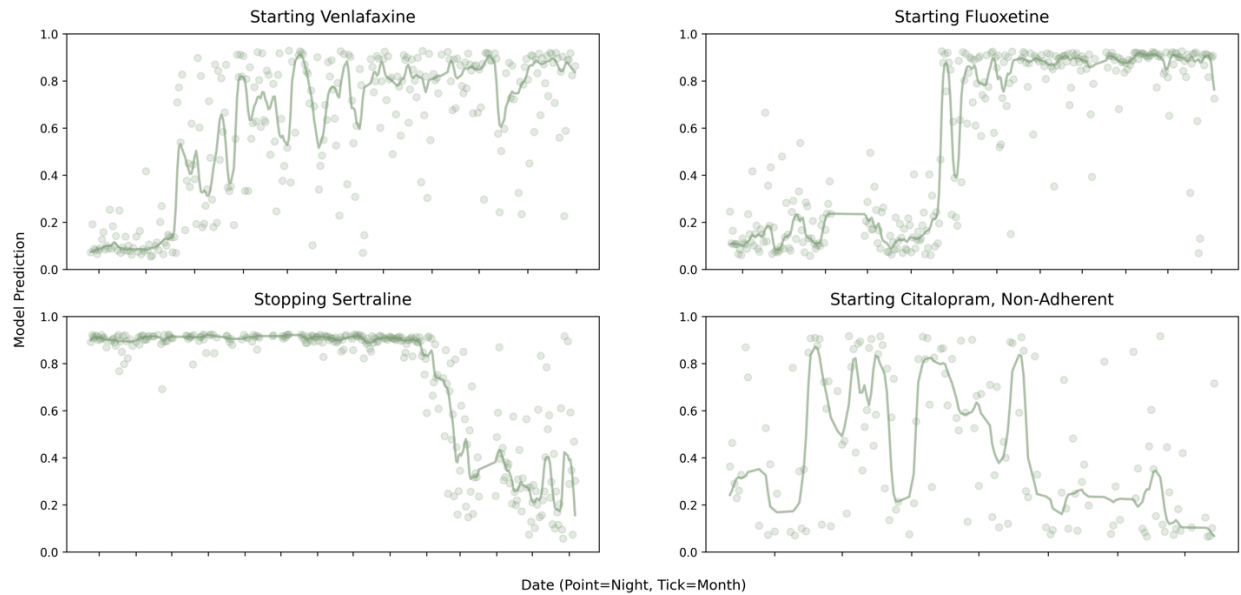

**Figure 5 | Longitudinal tracking of antidepressants using the adherence biomarker.** *Four individuals from the MIT dataset were monitored over several months, each undergoing a change in antidepressant treatment. Each dot represents a single night of sleep; x-axis ticks denote calendar months. Lines represent smoothed trajectories of nightly biomarker prediction scores. **a,** Patient who initiated venlafaxine. **b,** Patient who initiated fluoxetine. **c,** Patient who discontinued sertraline. **d,** Patient who initiated fluoxetine but exhibited irregular adherence.*

## Model Interpretability

To gain insight into how the model distinguishes antidepressant users from controls, we examined its latent space and reconstructed EEG representations. A t-SNE projection (Fig. 6a) of the model's latent space (i.e., classification token before the projection head) revealed that samples tended to cluster in a manner consistent with REM latency, with longer REM latency generally associated with higher model prediction scores (i.e., increased likelihood of taking antidepressant). Comparing the distribution of the model classification scores with the distribution of REM latency in the t-SNE space (Fig 6a left vs. right) confirmed this trend, and a Pearson correlation analysis (Fig. 6b) demonstrated a significant association between model output and REM latency (r = 0.439, p<1e-150) within the antidepressant cohort. These findings suggest that the model computes an internal representation analogous to REM latency, leveraging it as one source of evidence for antidepressant classification. However, the modest magnitude of the correlation and the incomplete alignment in the t-SNE space indicate that the model also relies on additional more discriminative features in its decisions.

To explore these additional features, we examined the model's intermediate EEG reconstructions derived from respiration data. We computed the reconstructed EEG power spectra for each night and compared spectral differences between antidepressant and control cohorts (Fig. 6c). Antidepressant users exhibited higher power in reconstructed EEG activity relative to the control cohort in the slow oscillation (0–1 Hz) and beta (16–32 Hz) frequency bands (all p < 1e-100). The 95% confidence intervals showed minimal overlap between groups, indicating a robust and separable spectral signature associated with antidepressant use. Interestingly, such differences in EEG slow oscillation and beta power are consistent with the antidepressant literature[10,54,55,56,57]. Supplemental Figure C extends this analysis to explore the correlation between EEG powers in early sleep and subsequent REM latency, pointing to a possible connection of the two. Overall, these results demonstrate that the model's reconstruction of EEG signals provides a richer set of discriminative features than REM latency alone, supporting the notion that the biomarker reflects a multifaceted neurophysiological profile of antidepressant effects.

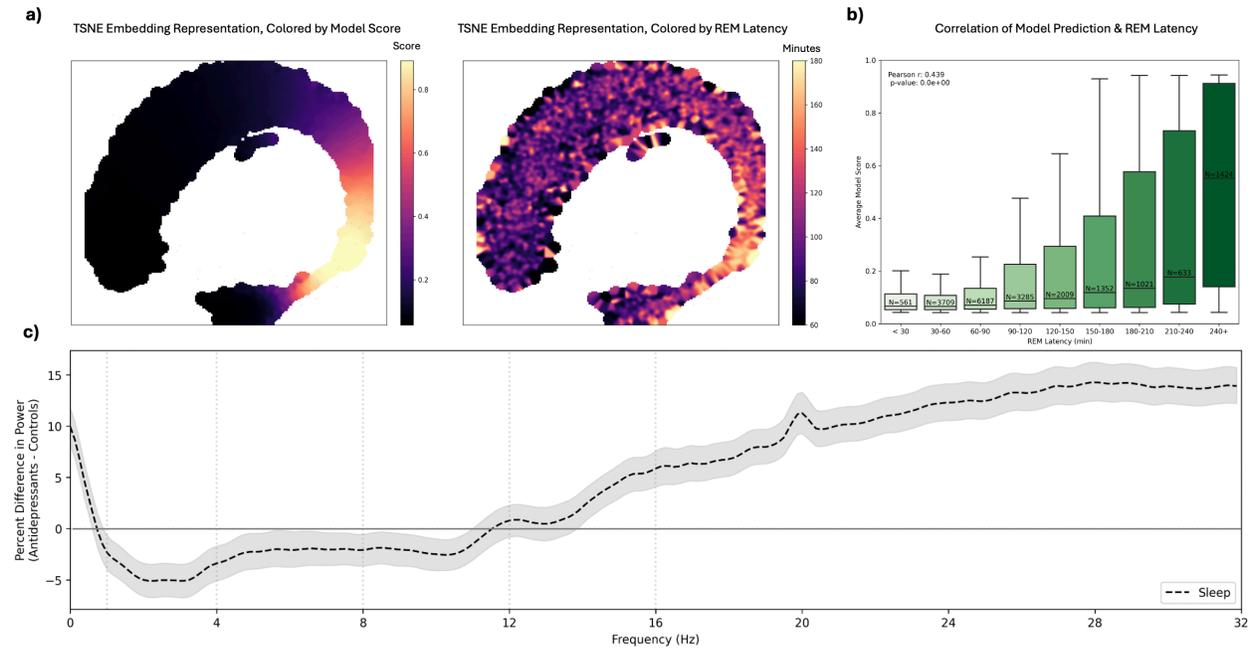

**Figure 6 | Model interpretability reveals REM latency associations and distinct EEG spectral signatures of antidepressant use**. *a,* Two-dimensional t-SNE projection of the model's latent space colored by model score (left) and REM latency (right), showing alignment between longer REM latency and model prediction of antidepressant use. *b,* Correlation of model score and REM latency reveals a significant positive correlation (r = 0.439, p < 1e-150). *c,* Differences in model-reconstructed EEG power spectra between the antidepressant and cohorts. Power is first averaged across time then across individuals within a cohort, then the difference is taken between the two cohorts, then normalized. The more the deviation from zero, the larger the difference between the cohorts. On average, antidepressant users exhibit pronounced alterations in slow oscillations (0–1 Hz) and beta activity (16–32 Hz), with minimal overlap of 95% confidence intervals, indicating a robust, separable neurophysiological signature. The model learned EEG reconstruction shows strong deviation from zero in those bands, allowing it to separate antidepressant users from controls.

# DISCUSSION

We present, to our knowledge, the first noninvasive, at-home biomarker that detects antidepressant intake directly rather than relying on proxies such as pill counts or pharmacy records. We validated the biomarker (AUROC = 0.84) on a large dataset of more than 20,000 participants, including 1,800 antidepressant users. Performance generalizes across antidepressant classes and is particularly strong for the widely used SNRI and SSRI categories. The biomarker is dose-sensitive, with scores increasing with dosage. Importantly, it demonstrates robustness to confounding from commonly co-prescribed psychiatric medications, including benzodiazepines and anticonvulsants. Its ability to isolate antidepressant use relative to other psychotropic medications, including anticholinergics, whose activity spans many medication classes, is an important advantage, as these drugs are often co-prescribed with antidepressants and contribute to polypharmacy, a known risk for nonadherence[42,58]. Polypharmacy has also been linked to greater difficulty in assessing medication adherence[59]. The biomarker further demonstrated robustness in detecting SSRIs and SNRIs with respect to the presence and severity of sleep apnea, which is important given SA's high prevalence and its impact on depression, including the severity of depressive symptoms[51,52].

The biomarker is enabled by a novel AI model that first reconstructs a representation of the patient's sleep EEG from nocturnal breathing, an easily accessible signal via consumer wearables or even contact-free sensors. A classification head then takes the reconstructed EEG and learns to distinguish antidepressant users from controls. In all settings, the model is trained on single nights of data, enabling independent, per-night decisions. This, in turn, supports continuous monitoring and extraction of trends associated with medication changes and nonadherence.

A key feature of this biomarker is that it can be assessed remotely and daily in the patient's home without active patient involvement. A contact-free sensor, akin to a Wi-Fi router (e.g., the Emerald sensor), placed in the bedroom captures nocturnal breathing and feeds it to the AI model to assess antidepressant use for that day; results are uploaded to a portal accessible to the physician. This approach addresses key gaps in antidepressant adherence monitoring: it is cost-effective, requires no clinic visit or active patient engagement, and infers ingestion from physiological effects rather than administrative surrogates (e.g., pharmacy fills, pill counts, or electronic logs), which have significant limitations[59,60]. Because the assessment is passive and remote, it can overcome adherence barriers observed in some populations[61,62], including many older adults and individuals with unipolar depression. Both unintentional nonadherence (e.g., forgetfulness) and intentional nonadherence (e.g., concerns about side effects, stigma, cost, or perceived lack of efficacy) are common and difficult to detect with conventional tools. Timely, patient-specific adherence information could enable clinicians to adjust treatment proactively and reduce risks associated with abrupt discontinuation, including withdrawal, relapse, and suicidal ideation [1,4,5].

The findings also have implications for clinical research. Unmeasured adherence is a major source of bias in trials evaluating treatments[63], including those for major depressive disorder (MDD); psychotherapy trials (e.g., CBT) are further complicated by concurrent antidepressant use[64]. The proposed biomarker provides an objective, quantitative method to verify adherence, helping ensure that pharmacologic exposure is properly accounted for before crossover phases and improving the interpretability of clinical outcomes.

Several limitations merit consideration. First, the current model yields a binary indicator of antidepressant use. Adherence, however, is more nuanced than ingestion alone and includes whether the patient takes the correct dose at the appropriate time of day. Although the present model does not capture these dimensions, it remains valuable for identifying and managing nonadherence. Second, we investigated potential confounders, including age, sex, SA, and various psychotropic medications, and observed robustness; nonetheless, other drugs or conditions may still confound performance. Third, underlying conditions commonly treated with antidepressants (e.g., depression and anxiety) themselves alter sleep; incomplete phenotyping limits our ability to disentangle antidepressant pharmacologic effects from disease- and treatment-related effects.

These constraints point to clear directions for future work. Prospective studies with ground-truth adherence, though difficult to establish with current measurement methods, will be essential to quantify accuracy and temporal resolution in real-world settings. Model extensions that encode pharmacologic specificity (individual medications, dose, time of day, pharmacokinetic profiles, and interactions with all co-prescribed agents) are a natural next step. A systematic evaluation of dose–response relationships between antidepressant exposure and biomarker scores could inform individualized dosing strategies, including early in treatment before emergence of full therapeutic effects when patients are most vulnerable to nonadherence and treatment discontinuation as a result of perceived lack of benefits and early adverse effects [65] [66] [67]; optimizing therapy at this stage may minimize side effects and improve the success of therapy[40]. Finally, prospective studies that capture both medication use and clinical outcomes will allow us to test whether nocturnal respiration patterns serve as a continuous physiological correlate of treatment response or remission.

In summary, our results show that sleep-related physiological changes can be distilled into a reliable, scalable biomarker of antidepressant use via AI analysis of nocturnal respiration. Moreover, this can be achieved with minimal patient burden, in a contactless, remote manner, offering significant benefits for both patients and clinicians involved in monitoring therapy. This advances digital psychiatry and pharmacotherapy by introducing a contactless, physiology-driven tool for antidepressant adherence monitoring and lays the groundwork for remote, personalized treatment strategies in MDD and related disorders.

# ETHICS DECLARATIONS

D.M. has received research support from Nordic Naturals and Heckel Medizintechnik GmbH and honoraria for speaking from the Massachusetts General Hospital (MGH) Psychiatry Academy. He collaborates with the MGH Clinical Trials Network and Institute, which receives research funding from multiple pharmaceutical companies and the U.S. National Institute of Mental Health (NIMH). D.K. receives research funding from the NIH, NSF, Sanofi, Takada, IBM, Gwangju Institute of Science and Technology, Wistron, KACST, Michael J Fox Foundation, Helmsley Charitable Trust, and the Rett Syndrome Research Trust, is a cofounder of Emerald Innovations, Inc., and serves on the scientific advisory board of Graviton Therapeutics and the board of directors of Cyclerion Inc. M.F. has received research support over the past three years from numerous pharmaceutical companies, foundations, and federal agencies, including Abbvie, Janssen, Pfizer, Takeda, the NIH, and PCORI. He holds equity in Neuromity, Psy Therapeutics, and Sensorium Therapeutics, and has licensed patents for the Sequential Parallel Comparison Design (SPCD), pharmacogenomics of depression treatment, and other innovations through MGH. His consulting activities have been conducted solely through MGH, except for past roles with Revival Therapeutics and Sensorium Therapeutics, both no longer in existence. He is also the copyright holder of several clinical assessment tools. All other authors declare no competing interests.

# SUPPLEMENTAL MATERIAL

Additional Information about Data

**Respiration Signal Preprocessing:** Respiration signals were high-pass filtered, resampled, and normalized. Belt data were sampled at 10 Hz, and wireless signals were processed into a filtered respiration signal using a standard algorithm adapted from Yue et al[17].

**EEG Signal Preprocessing:** In datasets with full polysomnography, EEG recordings were used during model training but not during inference. Signals were extracted from the C4–A1 electrode pair, band-pass filtered, and resampled to 64 Hz. To mitigate inter-site and device heterogeneity, EEG amplitudes were rescaled for each dataset using fixed factors derived from age- and sex-matched cohorts. Signals were then converted into multi-taper spectrograms using a 30-second sliding window and a frequency range capped at 32 Hz.

**Quality Control and Exclusion Criteria:** Nights with fewer than four hours of recorded sleep were excluded from all analyses (which amounted to 3.5% of all night and 2.9% of nights with antidepressants). This threshold was selected to allow for a minimum of two sleep cycles, avoid naps for in-home data and bad sensor data for sleep-lab studies. No imputation was performed; only nights with complete respiration and label data were retained.

**Label Control:** All patients with reported antidepressant use were assumed to be adherent for model training and evaluation. Naturally, different studies have different criteria for assessing medication use as part of study protocol. Details regarding the manner of collection of medication information in each dataset used in this study are provided below:

> **WSC** – Medication information was collected through patient self-report. Before completing the sleep study, patients filled out a questionnaire regarding medication use, and the following was the phrasing of the question:
>
> "Do you regularly take any medicines? If yes, please list the name of each drug"
>
> **CFS** – Medication information was collected through patient self-report. Before completing the sleep study, patients filled out a questionnaire regarding medication use, and the following was the phrasing of the question:
>
> "74b. Regular medicine: Mark if used in the last three days", with choices for: Antidepressant, Prozac, Zoloft, Paxil, Celexa, Other Antidepressants
>
> **MIT** – Patients visited the physician at Baseline, 6 months, 12 months, and depending on the duration of the study, at 24 months. During each encounter, a list of active medication was updated, including new medications and discontinued medications, with a patient estimate of the date of the medication change.
>
> **SHHS** – Participant taking antidepressant within *two weeks* of the Sleep Heart Health Study Visit One (SHHS1) visit. All medications were recorded during the interview (medications were physically shown to the physician), and medication information was later categorized by physician review.
>
> **MROS** – Participants were asked to bring all prescription and nonprescription medications used *within the 30 days* preceding the clinic visit. If a participant forgot to bring one or more medications, clinic staff was responsible for obtaining this information over the telephone or at a return visit. All medications were entered into an electronic database, verified by pill bottle examination, and each medication was matched to its ingredient(s) based on the Iowa Drug Information Service (IDIS) Drug Vocabulary

**HCHS** – HCHS/SOL records all prescription and over-the-counter medications used by participants in the *past four weeks*. If a participant forgot to bring one or more medications, clinic staff was responsible for obtaining this information over the telephone or at a return visit.

## Analysis of Drug Co-Therapy as a Confound for Biomarker Performance

To evaluate biomarker specificity, we examined drug co-therapy effects. Datasets were restricted to those with exact medication annotations, as in Figure 3d. Antidepressant users, classified as SSRI, SNRI, or TCA, were stratified into three groups: (i) single antidepressant use, (ii) antidepressant plus benzodiazepine co-therapy, and (iii) multi-antidepressant use, defined as concurrent prescriptions for more than one antidepressant. The score distribution for each group is shown in Figure A, with the control cohort included for reference. Co-therapy groups exhibited distributions comparable to single antidepressant users, indicating that co-therapy does not confound model performance. Notably, the multi-antidepressant group demonstrated higher scores relative to single antidepressant use. The TCA antidepressant medication cohort exhibited significant improvement in model score with drug co-therapy.

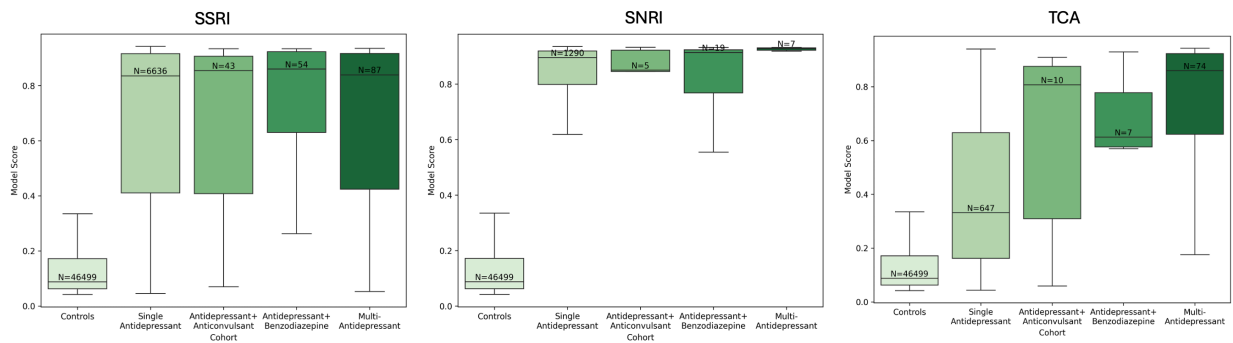

**Supplementary Figure A | Impact of drug co-therapy on the biomarker score for SSRI, SNRI, and TCA. The control cohort is included in the plots for reference.**

## Analysis of Obstructive Sleep Apnea as a Confound for Biomarker Performance

We assessed the biomarker's performance across varying levels of sleep apnea severity in three major antidepressant classes. Data were restricted to cohorts with expert-labeled apnea events and apnea–hypopnea index (AHI) calculations. The AHI was determined following AASM clinical guidelines. Participants were classified as Normal (AHI < 5), Mild (AHI 5–15), Moderate (AHI 15–30), or Severe (AHI > 30). Model score distributions were then compared between antidepressant users and controls, as shown in Fig. B. While SSRIs and SNRIs demonstrated non-significant changes to model score with increasing apnea severity, model score for TCA decreased with mild to moderate sleep which could be due to the sedative effects of TCAs being counteracted by the increased arousal associated with sleep apnea. Importantly, both mild and moderate groups maintained significantly higher model scores compared to controls ($p < 1\text{e-}200$).

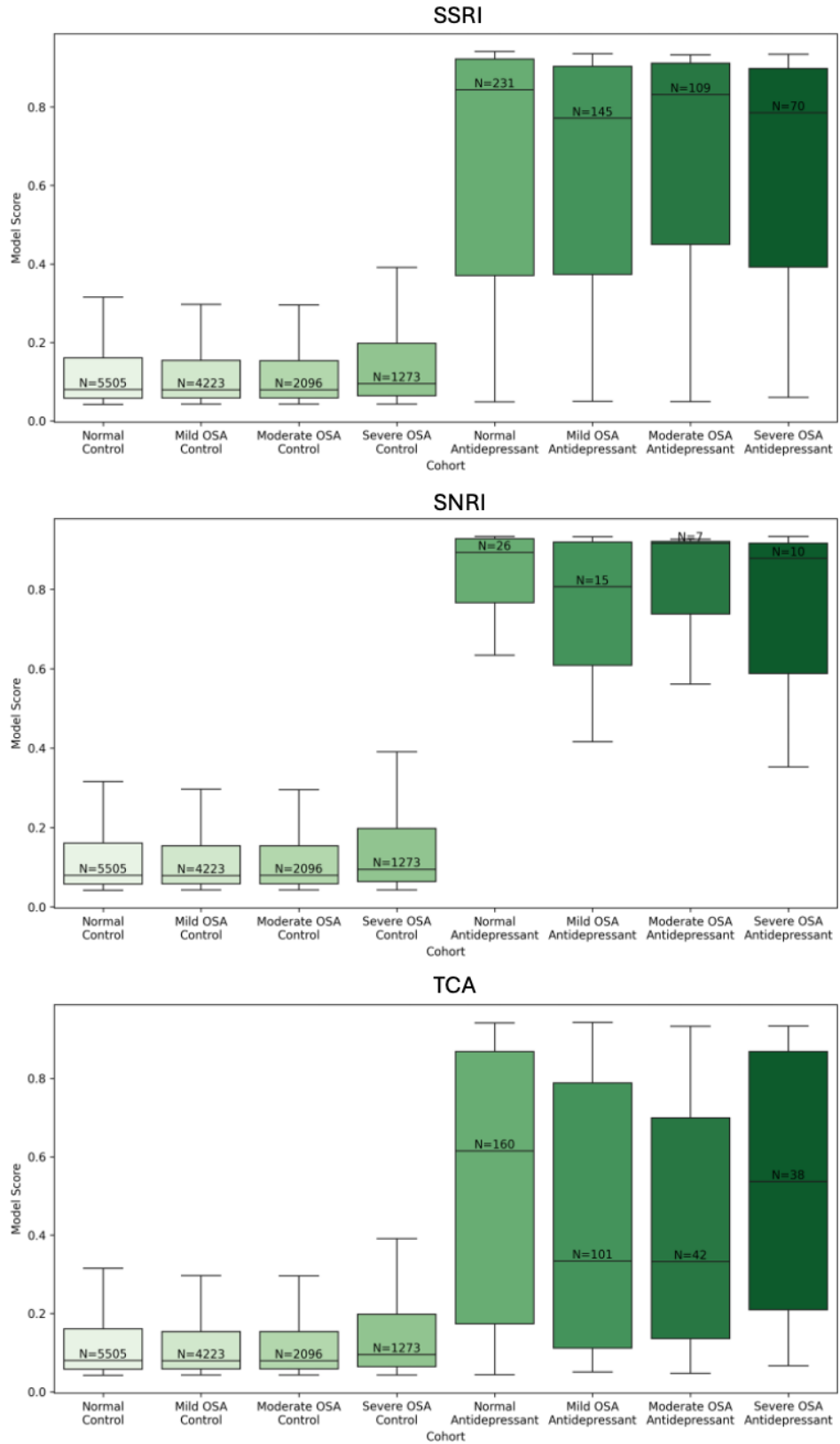

**Supplementary Figure B | Impact of sleep apnea on biomarker score for SSRI, SNRI, and TCA.**

# Correlation of REM Latency with Slow Oscillation and Beta Power in Early Sleep

To further examine the relationship between REM latency and model-predicted EEG power estimates (Fig. 5), we directly compared the predicted slow oscillation (SO) and beta power with individual REM latency. We focus on sleep preceding REM onset, as high power prior to REM onset could increase REM latency. The mean band power was calculated in early sleep, defined as the first hour of NREM sleep. Analyses were conducted separately for the antidepressant and control cohorts. In the antidepressant cohort, SO power correlated significantly with REM latency ($r = 0.20$), as did beta power ($r = 0.28$). When SO and beta powers were summed, the correlation increased ($r = 0.35$). In the control cohort, SO power did not significantly correlate with REM latency ($r = 0.03$), whereas beta power did ($r = 0.24$), with a summed power correlation of $r = 0.21$.

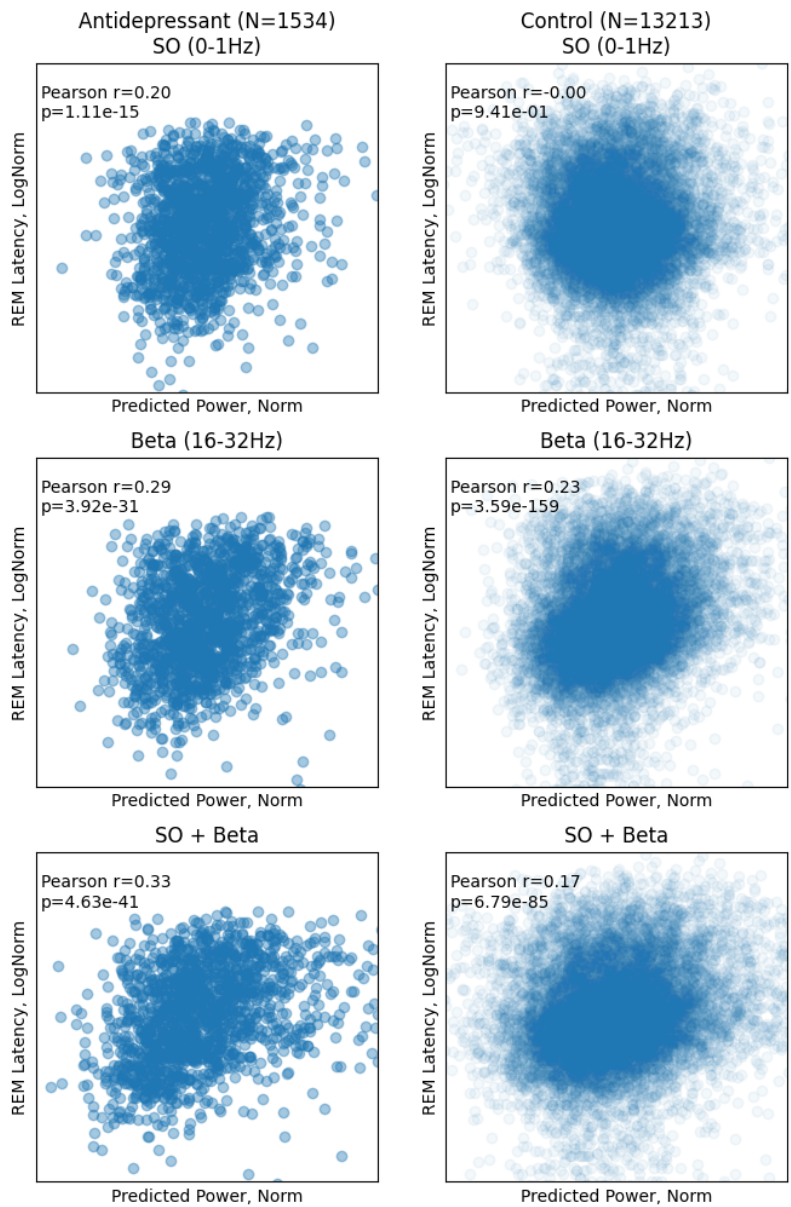

**Supplementary Figure C | Correlation between REM latency and the power of slow waves (SO) and beta band in the reconstructed EEGs in the first hour of sleep prior to the first REM cycle.**

## Antidepressant Dosing Normalization

To compare biomarker performance across different antidepressant dosages, individual prescriptions were normalized using the ratio of Prescribed Daily Dose (PDD) to Defined Daily Dose (DDD). DDD values for each antidepressant medication were obtained from the World Health Organization Drug Statistics Database [68] and are summarized in the accompanying table.

| Antidepressant | Defined Daily Dose (mg) |
|---|---|
| Escitalopram | 10mg |
| Citalopram | 20mg |
| Fluoxetine | 20mg |
| Sertraline | 50mg |
| Paroxetine | 20mg |
| Mirtazapine | 30mg |
| Bupropion | 300mg |
| Venlafaxine | 150mg |
| Desvenlafaxine | 50mg |
| Imipramine | 100mg |
| Nortriptyline | 75mg |

**Supplementary Figure D | Prescribed Daily Dose (PDD) according to the World Health Organization Drug Statistics database.**

## Random Forest Baseline Description

Two random forest baseline models were implemented to compare our deep learning model to baselines that operate on sleep-stage metrics known to be impacted by antidepressant use. The first model included sleep architecture features, such as stage durations, stage latencies, number of awakenings, and stage transition counts. The second model incorporated the same metrics augmented with sleep EEG power–based features, including absolute and relative band powers, per-stage band powers, and band power sleep ratios. Model confidence intervals were estimated via bootstrapping. The HCHS and MIT datasets were excluded from these analyses owing to the absence of ground-truth sleep stage annotations.

| Stage Metrics | | | | Additional EEG Metrics | | |
|---|---|---|---|---|---|---|
| **Metric Category** | **Specific Metrics** | **Total Metrics** | | **Metric Category** | **Specific Metrics** | **Total Metrics** |
| Stage Durations | Wake Duration, N1 Duration, N2 Duration, N3 Duration, REM Duration, Total Sleep Duration | 6 | | Band Powers | SO, Delta, Theta, Alpha, Sigma, Beta I, Beta II | 7 |
| Sleep Latencies | Sleep Onset Latency, REM Onset Latency | 2 | | Stage Band Powers | SO, Delta, Theta, Alpha, Sigma, Beta I, Beta II x Wake N1 N2 N3 REM | 35 |
| Awakening | WASO, Number of Awakenings, Sleep Efficiency | 3 | | Sleep Ratio (Cycle 1 vs) | SO, Delta, Theta, Alpha, Sigma, Beta I, Beta II x N2 N3 REM | 21 |
| Sleep Stage Transition Counts | W/N1/N2/N3/REM 5x5 | 25 | | | | |
| Demographics | Age, Sex | 2 | | | | |

**Supplementary Figure E | Features used in the baseline models**

# TSNE Representations of Biomarker, All Folds

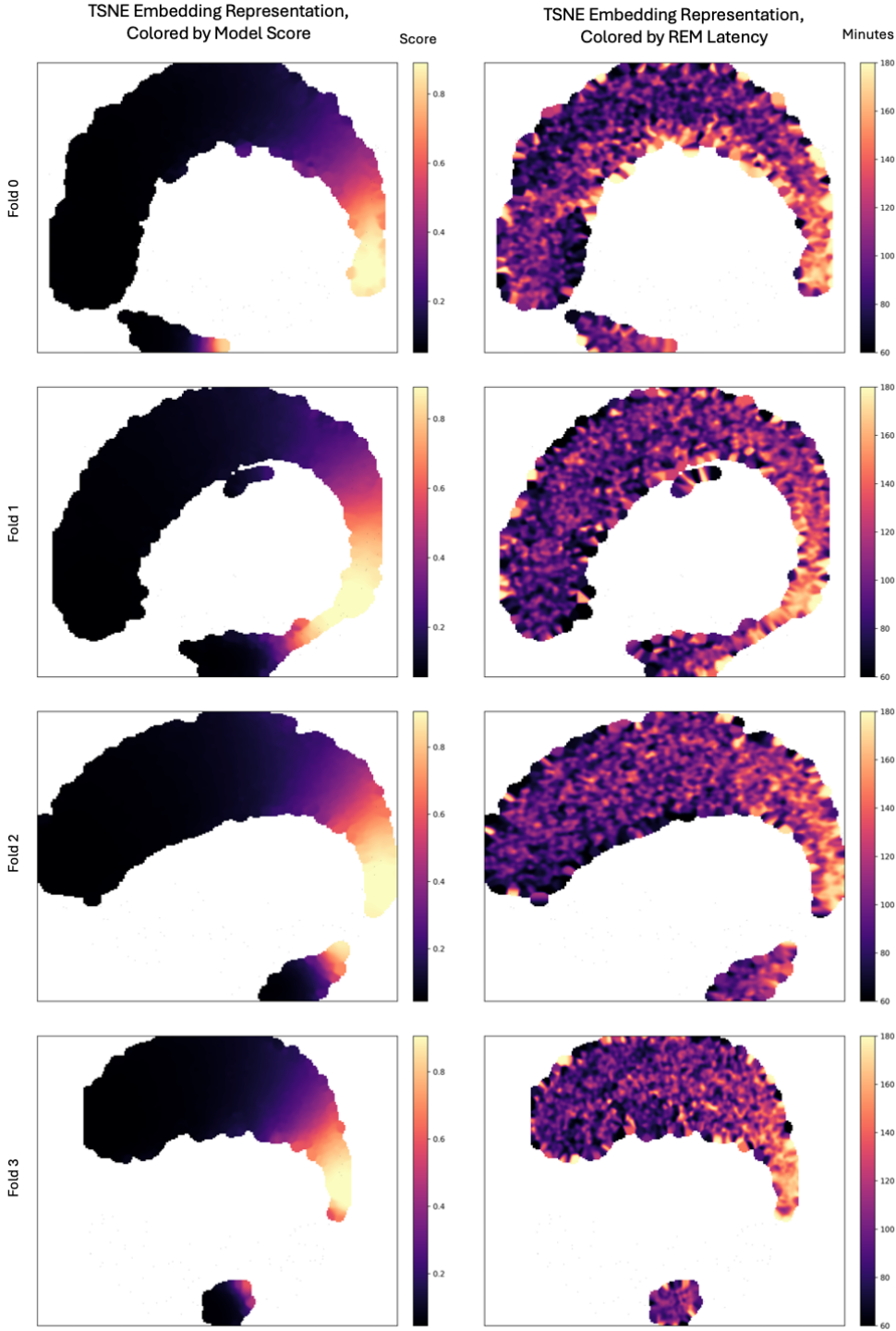

**Supplementary Figure F |** Figure 5a illustrates the t-SNE representation of biomarker embeddings from one fold of the dataset. To confirm robustness, we visualized all four folds of the data and corresponding models, which demonstrated consistent alignment between model predictions and REM latency across folds.